\documentclass[letterpaper, 10 pt, conference]{ieeeconf}
\usepackage{subcaption}
\usepackage{amsmath} 
\usepackage{amssymb}  
\usepackage{amsfonts}
\usepackage{hyperref}

\usepackage{amsthm}
\usepackage{algorithmic}
\usepackage{graphicx}
\usepackage{textcomp}
\usepackage{xcolor}
\usepackage{cite}
\usepackage[compact]{titlesec}
\newtheorem{theorem}{Theorem}

\IEEEoverridecommandlockouts                          

\title{\LARGE \bf
Delay-aware Robust Control for Safe Autonomous Driving}
\begin{document}


\author{Dvij Kalaria$^{1}$, Qin Lin$^{2*}$, and John M. Dolan$^{3}$
\thanks{$^{1}$Dvij Kalaria is with the Department of Computer Science and Engineering, IIT Kharagpur, India {\tt\small dvij@iitkgp.ac.in}}%
\thanks{$^{2}$Qin Lin is with the Electrical Engineering and Computer Science Department, Cleveland State University {\tt\small q.lin80@csuohio.edu} (*corresponding author)}%
\thanks{$^{3}$John M. Dolan is with the Robotics Institute, Carnegie Mellon University {\tt\small jdolan@andrew.cmu.edu}}
}

\maketitle

\begin{abstract}
With the advancement of affordable self-driving vehicles using complicated nonlinear optimization but limited computation resources, computation time becomes a matter of concern. Other factors such as actuator dynamics and actuator command processing cost also unavoidably cause delays. In high-speed scenarios, these delays are critical to the safety of a vehicle. Recent works consider these delays individually, but none unifies them all in the context of autonomous driving. Moreover, recent works inappropriately consider computation time as a constant or a large upper bound, which makes the control either less responsive or over-conservative. To deal with all these delays, we present a unified framework by 1) modeling actuation dynamics,  2) using robust tube model predictive control, and 3) using a novel adaptive Kalman filter without assuming a known process model and noise covariance, which makes the controller safe while minimizing conservativeness. On the one hand, our approach can serve as a standalone controller; on the other hand, our approach provides a safety guard for a high-level controller, which assumes no delay. This can be used for compensating the sim-to-real gap when deploying a black-box learning-enabled controller trained in a simplistic environment without considering delays for practical vehicle systems.

\end{abstract}

\section{Introduction}
The recent surge in autonomous driving has led to an increase in demand for increased affordability and accessibility. The stringent requirements of onboard computing and high-resolution sensors pose a major challenge in meeting this demand. While there has been much work on making algorithms more efficient, the required computation time is still dangerous in fast-changing environments such as high-speed scenarios. Most commonly used control algorithms assume the planned command to be applied instantaneously. This assumption can generate significant tracking errors and jeopardize stability.


 
This work presents a unified approach to dealing with three types of delay: 1) computation time delay; 2) actuator command processing delay; and 3) actuator dynamics delay. First, we model the actuator's steering dynamics delay using a first-order ordinary differential equation (ODE). Second, we propose a delay-aware robust tube Model Predictive Control (MPC). It is coupled with our proposed filter, called INFLUENCE - adaptIve kalmaN FiLter with Unknown process modEl and Noise CovariancE (including process noise and measurement noise). INFLUENCE is a novel adaptive Kalman filter variant combining online identification of an unknown dynamic model and estimation of noise covariances. INFLUENCE probabilistically safely estimates the computation time delay. We propose two controller plans: the first plan serves as a standalone controller for a delay-aware robust control; in the second controller plan, we compensate a learning-enabled (LE) primary controller to boost its safety performance. This controller plan has application merit, since LE controllers are being used in autonomous systems. However, simplistic assumptions are usually made in the training procedure of the LE controllers. For safety-critical systems such as autonomous driving, it is crucial to close the sim-to-real gap by providing a safety-assured guard. We treat the LE controller as a high-level controller. In the low-level control part, we track the reference generated by the LE controller, but actively regulate its unsafe control by delay compensation.

In summary, we make the following novel contributions:

1. A unified delay-aware robust control approach dealing with three major delays: computation time delay, actuator command processing delay, and actuator dynamics delay.

2. INFLUENCE, a probabilistic framework for real-time estimation of computation time instead of taking an upper bound, which makes the algorithm safe while minimizing conservativeness. INFLUENCE has application merit for general safe prediction problems, since it does not assume known process model and noise covariance.

3. A control plan to safely compensate for these delays for a LE controller which doesn't consider delays.

The rest of this paper is organized as follows. Section \ref{sec:related_work} provides a review of some important related work. Section \ref{sec:method} describes the methodology. Section \ref{sec:experiment} presents the experimental results. The conclusions are in Section \ref{sec:conclusions}.

\section{Related works}
\label{sec:related_work}


In the low-level control layer of a vehicle system, many algorithms ignore the delay arising from various factors such as computation, actuator command processing, sensor delay, and actuator dynamics. For the compensation of computation delay for discrete MPC, \cite{cortes2011delay} proposes the simple solution of shifting the initial state by one control step, approximating it from the prediction model. However, this is not suitable if the computation time is more than one control step. \cite{su2013computation} further proposes to use a buffer to store control commands from the previous batch. It also proposes the use of a pre-compensating unit and robust tube MPC to prove the safety of the system under bounded perturbation. This plan is well suited for static scenarios, where the objective is nearly constant throughout. However, for highly dynamic scenarios where the planned trajectory changes rapidly, taking the upper bound as the horizon length might lead to the algorithm's being less responsive. We instead propose to use an adaptive Kalman filter to approximate a local upper bound on computation time and adapt to its changing values instead of taking an upper bound as the horizon length. 

For compensation of delay due to actuator dynamics, \cite{nahidi2019study} proposes approximating the actuator dynamics by a first-order ODE, after which the actuator state can be augmented in the state space model. The approach has been tested on differential braking stability control. Instead, we use a similar design for compensating steering delay.

For considering delay caused by processing of actuator commands at the actuator, \cite{liao2018design} proposes an initial state transition method similar to the compensation for computation delay, but a preview continuous controller. It proposes a closed-loop solution to compensate for a dead time between when the command is planned and when it reaches the actuator. \cite{6083022} further extends the idea by including compensation for actuator saturation, as well to make the solution deployable on real systems with control limits. However, with the use of the preview controller, it becomes difficult to include state constraints in the system. \cite{10.1115/DSCC2014-6269} proposes a simple way to compensate for the sensor delay by transforming the frame of sensor values to reflect values at the current time and not at the time when they were recorded. Our work, however, considers computation delay, actuator command processing delay, and actuator dynamics as well as control and state constraints under one optimization framework in the context of autonomous vehicle control.

\section{Methodology}
\label{sec:method}

\subsection{Notation}
A polytope is defined as a convex hull of finite points in $d$-dimensional space $\mathbb{R}^d$.
The Minkowski sum of two polytopes is defined
as $P \oplus Q := \{x + q \in \mathbb{R}^d : x \in P, q \in Q\}$.
The Pontryagin difference of two polytopes is defined as $P \ominus Q := \{x \in \mathbb{R}^d : x + q \in P, q \in Q\}$.

\subsection{System dynamics} \label{actuator_dynamics}

A kinematic bicycle model describes the dynamics of the vehicle with the state variables being position ($p_x$, $p_y$), heading angle ($\theta$), and velocity ($v$), and the control variables acceleration ($a$) and steering angle ($\delta$). It is commonly assumed in the literature that steering angle $\delta$ and acceleration $a$ are applied instantaneously by the actuators. But in reality, there is a certain lag between the command and when the actuator physically modifies the steering angle state, which is called the actuator dynamic delay. We modify the system dynamics to include the actual steering angle as a state denoted as $\delta_{a}$. The control command is now the desired steering angle $\delta$. We approximate the change in the steering angle state by a first-order ODE similar to \cite{nahidi2019study}, i.e., $\Dot{\delta_a} = K_{\delta} (\delta - \delta_a)$, where $K_{\delta}$ is the inverse of the time constant for the steering actuator. For acceleration, pedal dynamics are assumed to be instantaneous for the experiments in this paper. However, they can also be approximated in the same way. After discretization, the vehicle state is now modified as $x_k = [p_{x,k},p_{y,k},\theta_k,v_k,\delta_{a,k}]$. The discrete dynamics are given in Eq. \ref{dyn_eqn_2}.

\begin{subequations} \label{dyn_eqn_2}
    \begin{flalign}
        &p_{x,k+1} = p_{x,k} + \frac{\sin (\theta_k + \kappa_k l_k) - \sin(\theta_k)}{\kappa_k} \\
        &p_{y,k+1} = p_{y,k} + \frac{\cos (\theta_k) - \cos(\theta_k  + \kappa_k l_k)}{\kappa_k} \\
        &\theta_{k+1} = \theta_{k} + \kappa_k l_k\\
        &v_{k+1} = v_k + a_k \Delta t \\
        &\delta_{a,k+1} = \delta_{k} - (\delta_{k} - \delta_{a,k}) (e^{K \Delta t} - 1)
    \end{flalign}
where the curvature $\kappa_k = \frac{\tan(\delta_{a,k}) C_r}{L}$, $L$ is the vehicle length, and the travel distance $l_k = v_k \Delta t + \frac{1}{2}a_k \Delta t^2$.
\end{subequations}

\subsection{Robust Tube MPC}

For a discrete linear system with system matrices $A\in \mathbb{R}^{n \times n}$ and $B\in \mathbb{R}^{n \times m}$, let the control gain $K \in \mathbb{R}^{m \times n}$ be such that the feedback system of $A_K = A + BK$ is stable. Let $\mathcal{Z}$ be the disturbance-invariant set for the controlled uncertain system $x^+ = A_Kx + w$, satisfying $A_K \mathcal{Z} \oplus \mathcal{W} \subseteq \mathcal{Z}$, where the disturbance $w$ is assumed to be bounded ($w \in \mathcal{W}$) by a polyhedron that contains the origin in its interior. The following finite optimization problem is solved at each step for $\Bar{X} = \{\Bar{x}_0,\Bar{x}_1...,\Bar{x}_N\}$, $\Bar{U} = \{\Bar{u}_0,\Bar{u}_1...,\Bar{u}_{N-1}\}$  and reference state sequence $X_{ref} = \{ x_{ref,0}, x_{ref,1}..., x_{ref,N} \}$ obtained from the path planner, where $N$ is the horizon length, and $\Bar{X}$ and $\Bar{U}$ are the state and control sequences of a nominal system ignoring $w$: 

\begin{equation} \label{objective function}
\begin{split}
    \mathop{\min}\limits_{\Bar{x}_0, U}&\quad\sum_{t=0}^{N-1} (\Bar{x}_k-{x_{ref,k}})^T Q (\Bar{x}_k-{x_{ref,k}})+\Bar{u}_k^T R \Bar{u}_k\\&+(\Bar{x}_N - {x_{ref,N}})^T Q_N (\Bar{x}_N - {x_{ref,N}})\\
    s.t. &\quad \Bar{x}_{k+1} = A \Bar{x}_k + B \Bar{u}_k\\
    &\quad x_{0} \in \Bar{x}_0 \oplus \mathcal{Z}\\
    &\quad \Bar{u} \in \mathcal{U} \ominus K\mathcal{Z}\\
    &\quad \Bar{x} \in \mathcal{X} \ominus \mathcal{Z}
\end{split}
\end{equation}
where $Q$, $R$ and $Q_N$ are the state, control and terminal state cost matrices, respectively. The control command given is $u = \Bar{u} + K(x-\Bar{x})$, where $x$ is the current state. This guarantees $x^+ \in \Bar{x}^+ + \mathcal{Z}$ for any $w \in \mathcal{W}$, i.e., all states $x_k$ will be inside the constraint set $\mathcal{X}$. However, for a nonlinear system as in our case, we use the equivalent LTV system, where the system matrices $A$ and $B$ are replaced with Jacobian matrices $A_k$ and $B_k$ at the current state for the system dynamics used in Eq. \ref{dyn_eqn_2}. For more details as well as a detailed proof of the feasibility and the stability of the above controller, see \cite{1383612}. Also, a mismatch between the linearized and the actual model can be compensated by adding an additional disturbance $\hat{w}$ assuming the model non-linear function to be a Lipschitz function \cite{gao2014}. For the experiments in this work, we assume the disturbance margin is large enough to cover this extra disturbance.

\subsection{Delay-aware robust tube MPC}
\begin{figure}[htbp]
    \centering
    \includegraphics[width=0.5\textwidth]{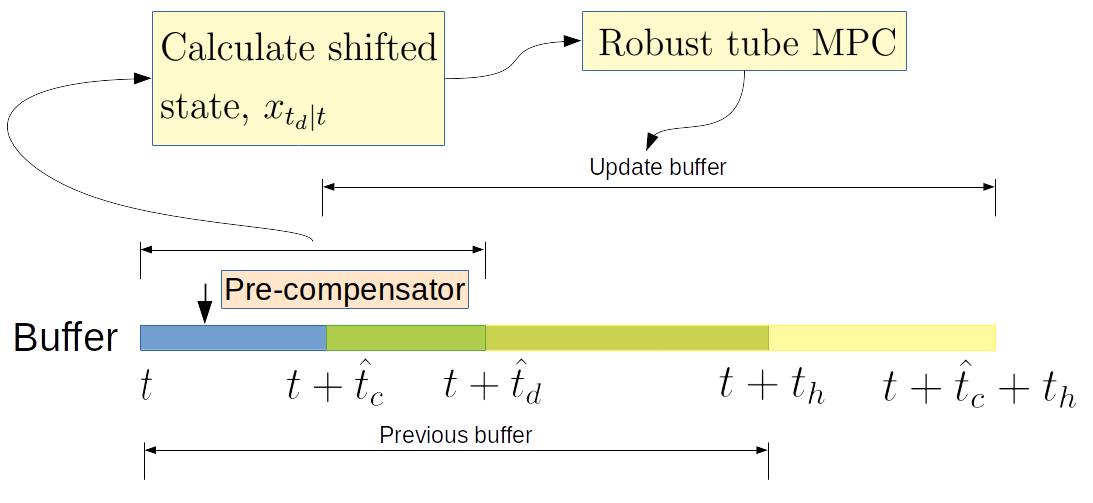}
    \caption{Dual cycle control scheme for tube MPC with delay}
    \label{dual_cycle}
\end{figure}

For the above formulation, we assumed delay time to be zero, meaning that the computed command is delivered to the system at the same time the observation is made for the current state $x$. But in practice, there is computation time denoted as $t_c$ and an actuator command processing delay $t_a$ after the calculated command is delivered, resulting in a total delay time $t_d = t_c + t_a$. Hence, if the current state $x$ is observed at time $t$, the computed command influences the actuator state at $t + t_d$ time. This may lead to instability of the system if $t_d$ is large and the robust tube assumptions no longer hold true. In order to tackle this problem, \cite{su2013computation} proposes a bi-level control scheme to deal with time delay and also proves robustness using the tube MPC. A buffer block of commands is used for communication between higher- and lower-level units, as depicted in Fig. \ref{dual_cycle}. 
At time $t$, the set of possible states at time $t + t_h$, where $t_h$ is the horizon length, is predicted. The high-level tube MPC updates the buffer with nominal states and control commands from time $t + t_h$ to $t + 2 t_h$. If the higher-level MPC requires a delay time $t_c$ that is less than $t_h$, the system waits for the remaining time $t_h - t_c$ for time alignment. However, we believe this is only suited for systems when the objective is nearly constant. For dynamically changing objectives as well as state constraints, it is necessary to update the reference path more frequently, since waiting for the full horizon path to be followed may lead to inconsistencies. In the case of autonomous driving, for dynamic scenarios where the reference path has to be updated frequently, it would not be feasible to wait for $t_h$ to get a new updated path. 

Hence, we propose to get a local probabilistic upper-bound estimate $\hat{t}_c$ of the computation time. We update the buffer from ($t+\hat{t}_c$ to $t+\hat{t}_c+t_h$) instead of from ($t+t_h$ to $t+2 t_h$), as shown in Fig. \ref{dual_cycle}. This increases the controller plan update rate for the higher-level MPC and also makes the controller robust to changing computation times. For estimation of the local upper bound $\hat{t}_c$, we use an adaptive Kalman filter, as further described in Section \ref{kalman_filter}. Considering $t_a$ to be the extra delay due to actuator command processing, we thus get a new local upper-bound delay time estimate $\hat{t}_d = \hat{t}_c + t_a$, which we use to find the initial state estimate $x_{\hat{t}_d|t}$ assuming no disturbance after $\hat{t}_d$ time given the current state $x_t$. It can be calculated by piece-wise integration of the system dynamics using the control commands from the buffer. Hence, a command issued at $t+\hat{t}_c$ will be executed at $t+\hat{t}_c+t_a = t+\hat{t}_d$, where we consider our initial state to be for optimization, as depicted in Fig. \ref{dual_cycle}. Mathematically, the updated objective function is described in Eq. \ref{objective function robust}, where $x_{\hat{t}_d+t}$ is the actual state after time $\hat{t}_d$. The calculated nominal discrete states ($\Bar{X}$) and controls ($\Bar{U}$) are used to fill the buffer B from time $t + \Hat{t}_c$ to $t + \Hat{t}_c + t_h$ as $\Bar{u}_{[t + \Hat{t}_c + k\Delta t, t + \Hat{t}_c + (k+1)\Delta t]} = \Bar{u}_k$ for $k \in \{0,1,2...N-1\}$, as shown in Fig. \ref{dual_cycle}. The pre-compensator unit is a low-level process which executes commands in the buffer at a higher frequency than the high-level MPC.

\begin{equation} \label{objective function robust}
\begin{split}
    \mathop{\min}\limits_{\Bar{x}_0, U}&\quad\sum_{k=0}^{N-1} (\Bar{x}_k-{x_{ref,k}})^T Q (\Bar{x}_k-{x_{ref,k}})+\Bar{u}_k^T R \Bar{u}_k\\&\quad \quad \quad +(\Bar{x}_N - {x_{ref,N}})^T Q_N (\Bar{x}_N - {x_{ref,N}})\\
    s.t. &\quad \Bar{x}_{k+1} = A \Bar{x}_k + B \Bar{u}_k\\
    &\quad x_{t_d+t} \in \Bar{x}_0 \oplus \mathcal{Z} \implies x_{t_d|t} \in \Bar{x}_0 \oplus \mathcal{Z} \ominus (\oplus_{j=0}^{s-1} A_K^j \mathcal{W})\\
    &\quad \ s = \left \lceil \frac{t_d}{\Delta t} \right  \rceil \\
    &\quad \Bar{u} \in \mathcal{U} \ominus K\mathcal{Z}\\
    &\quad \Bar{x} \in \mathcal{X} \ominus \mathcal{Z}
\end{split}
\end{equation}

\subsubsection{Control Constraints ($\mathcal{U}$)}

Limits on acceleration and steering are formulated as control constraints for the optimization problem.

\subsubsection{State Constraints ($\mathcal{X}$)}

For the state variables $\theta$ and $v$, we set the upper and lower bound for their range. However, for the state variables $p_x$ and $p_y$, free space is non-convex in nature, hence it becomes quite computationally expensive to set them in the non-convex form for the optimization problem. Hence we use the IRIS algorithm we have used in previous work \cite{khaitan2020safe} to derive a set of convex constraints which can be used for efficient optimization of the path tracking problem while also ensuring safety through collision avoidance. IRIS optimizes the objective of finding linear constraints for each obstacle such that the resultant convex space fits the largest possible ellipsoid. We set the seed for IRIS as the predicted position (without uncertainty) of the vehicle after time $\hat{t}_d$ to get the resultant convex space $\mathcal{X}$.

\subsubsection{Disturbance-invariant set ($\mathcal{Z}$)}

The disturbance-invariant set can be over-approximated using $\mathcal{Z}_k = \Sigma^{N}_{i=0} A_{K,k}^{i} \mathcal{W}$ \cite{article} where $A_{K,k} = A_k + B_k K_k$. However, in the presence of delay, $A_{K,k+1}$ would be different for the next control cycle, hence for robustness, $\mathcal{Z}_k$ must be sufficient to be covered by $\mathcal{Z}_{k+1}$.

\begin{theorem}
Given the optimization in Eq. \ref{objective function robust}, the disturbance-invariant set $\mathcal{Z}_k$ is calculated as $\mathcal{Z} = \bigcup\limits_{A_K \sim S_A} (\oplus_{j=0}^{N-1} A_K^j \mathcal{W})$, i.e., a union set with all possible values of initial heading angle, speed and steering angle within the admissible range, which determines the matrix $A_K = A + B K$. The invariant set guarantees robust initialization of the optimization problem. The calculation of $\mathcal{Z}$ is computationally expensive, hence is carried out offline. Calculating $\mathcal{Z}$ offline and using the fixed value online for calculating the safe region for the vehicle doesn't increase the computation time of the online algorithm significantly.
\end{theorem}

\begin{proof}
$x_{t_d|t_k}$ is the expected state assuming no disturbance after $t_d$ for the current time $t_k$, and the application of $\Bar{u}$ with feedback. $x_{t_d+t_k}$ is the actual state at $t_d+t_k$. We have $x_{t_d+t_k} \in x_{t_d|t_k} \oplus (\oplus_{j=0}^{s-1} A_k^j \mathcal{W})$, where $s = \left \lceil \frac{t_d}{\Delta t} \right \rceil$, From Eq. \ref{objective function robust}, we have $x_{t_d+t_k} \in \Bar{x}_{0|t_d+t_k} \oplus \mathcal{Z}$. Hence, in order to guarantee robustness, $x_{t_d|t_k} \in \Bar{x}_{0|t_d+t_k} \oplus \mathcal{Z} \ominus (\oplus_{j=0}^{s-1} A_K^j \mathcal{W})$, we need to ensure $\Bar{x}_{0|t_d+t_k}$ has a valid solution, i.e., $(\oplus_{j=0}^{s-1} A_K^j \mathcal{W}) \subseteq \mathcal{Z}$. As $s \leq N$ and $0_n \in \mathcal{W}$, we can establish 
\begin{equation}
    (\oplus_{j=0}^{s-1} A_K^j \mathcal{W}) \subseteq (\oplus_{j=0}^{s-1} A_K^j \mathcal{W}) \oplus (\oplus_{j=s}^{N-1} A_K^j \mathcal{W})
\end{equation}
Thus, $\oplus_{j=0}^{s-1} A_K^j \mathcal{W} \subseteq \oplus_{j=0}^{N-1} A_K^j \mathcal{W}$. $A_K$ is the Jacobian matrix, which depends on $\theta$, $v$, and $\delta$ of $x_{t_d|t_k}$. Let's define set $S_A = \{ A_K(\theta, v, \delta) | \theta \in [-\pi,+\pi], \delta \in [-\delta_{max},\delta_{max}], v \in [v_{min}, v_{max}]\}$, which consists of all possible matrices.
\begin{equation}
    (\oplus_{j=0}^{N-1} A_K^j \mathcal{W}) \subseteq \bigcup\limits_{A_K \sim S_A} (\oplus_{j=0}^{N-1} A_K^j \mathcal{W})
\end{equation}
Thus, $\oplus_{j=0}^{s-1} A_K^j \mathcal{W} \subseteq \bigcup\limits_{A_K \sim S_A} (\oplus_{j=0}^{N-1} A_K^j \mathcal{W})$. $\mathcal{Z}$ is chosen as $\bigcup\limits_{A_K \sim S_A} (\oplus_{j=0}^{N-1} A_K^j \mathcal{W})$, which concludes the proof.
\end{proof}
\subsection{Estimating computation time} \label{kalman_filter}
We propose INFLUENCE for estimating a local upper bound on computation time. The conventional Kalman filter faces a crucial challenge when the dynamic model and noise covariance are unknown. On the subject of adaptive filters addressing this challenge, existing approaches either assume an unknown dynamic model but known covariance \cite{liu2015safe} or unknown covariance but a known dynamic model  \cite{article_kalman,myers1976adaptive}. INFLUENCE assumes the process model $\gamma$, process noise variance $q$ and measurement noise variance $r$ are all unknown. Since measuring time is direct but noisy, it is reasonable to assume the measurement matrix to be the identity matrix. However, if the assumption does not hold in other applications, the identification can be facilitated using the similar approach for process model identification in the INFLUENCE algorithm. We assume both noise distributions to be Gaussian, independent and mutually uncorrelated throughout. To make the optimization tractable, INFLUENCE iteratively fixes $\gamma$ first to identify $q$ and $r$ (see Eq. \ref{eq:filter-c} to Eq. \ref{eq:filter-k}), which are then fixed to update $\gamma$ (see Eq. \ref{eq:filter-l} to Eq. \ref{eq:filter-m}).
For the identification of $q$ and $r$, an exponential moving average is maintained to estimate prediction error $e$ and measurement error $w$, respectively. Parameters $N_q$ and $N_r$ determine the influence of older values on the moving average: the greater their values, the greater the weighting of older values. Thus, an incremental update for $q$ and $r$ can be done in Eq. \ref{eq:filter-d} and Eq. \ref{eq:filter-j}. These update rules have been adapted from \cite{article_kalman,myers1976adaptive}, while the update rules for state transition parameters $\theta$ have been adapted from \cite{liu2015safe}. $F$ is the learning gain, and $\lambda = \frac{N_{\theta} - 1}{N_{\theta}}$ denotes the forgetting factor for estimation of $\theta$, where $N_{\theta}$ is the buffer size for the process model's identification. The initialization of INFLUENCE is in Eq. \ref{eq:filter-a}.  

\begin{subequations} \label{kalman_filter_eqn}
\begin{flalign}
&x_{0|0} = t_{c,0}, p_{0|0} = 0, q_0 = \epsilon, r_0 = \epsilon \nonumber \\ 
&e_0 = 0, w_0 = 0, F_0 = I_2, \gamma = [1 \ \ 0]^T \label{eq:filter-a} \\ 
&x_{n|n-1} = \gamma_{n-1}^T [x_{n-1|n-1} \ \ 1]^T \label{eq:filter-b} \\
&p_{n|n-1} = \gamma_{n-1,0}^2\ p_{n-1|n-1} + q_{n-1} \label{eq:filter-c} \\
&e_{n} = \frac{N_r-1}{N_r} e_{n-1} + \frac{1}{N_r} (t_{c,n}-x_{n|n-1})\label{eq:filter-d} \\
&\Delta r_n = \frac{((t_{c,n}-x_{n|n-1})-e_n)^2}{N_r-1} - \frac{p_{n|n-1}}{N_r}\label{eq:filter-e} \\
&r_n = \left \vert \frac{N_r-1}{N_r} r_{n-1} + \Delta r_n\right \vert \label{eq:filter-f} \\
&K_n = \frac{p_{n|n-1}}{p_{n|n-1} + r_n}\label{eq:filter-g} \\
&x_{n|n} = x_{n|n-1} + K_n (t_{c,n} - x_{n|n-1})\label{eq:filter-h} \\
&p_{n|n} = (1-K_n) p_{n|n-1}\label{eq:filter-i}\\
&w_{n} = \frac{N_q-1}{N_q} w_{n-1} + \frac{1}{N_q} (x_{n|n} - x_{n|n-1}) \label{eq:filter-j}\\
&\Delta q_n = \frac{p_{n|n}-\gamma_{n-1,0}^2\ p_{n-1|n-1}}{N_q} + \frac{((x_{n|n} - x_{n|n-1})-w_n)^2}{N_q - 1}\label{eq:filter-k}\\
&q_n = \left \vert \frac{N_q-1}{N_q}q_{n-1} + \Delta q_n \right \vert \label{eq:filter-l}\\
&\text{Let } \phi \text{ be } [x_{n-1|n-1} \ \  1]^T \nonumber\\
&F_n = \frac{1}{\lambda} \left (F_{n-1} - \frac{F_{n-1} \phi \phi^T F_{n-1}}{\lambda+\phi^T F_{n-1} \phi}\right) \label{eq:filter-m}\\
&\gamma_{n} = \gamma_{n-1} + F_{n} \phi (x_{n|n} - x_{n|n-1}) \label{eq:filter-n}
\end{flalign}
\end{subequations}
where $t_{c,n}$ is the observed computation time at step $n$, and $\gamma_{n-1,0}$ is the first element of the vector at step $n-1$.

For the local upper bound estimate, we use the predicted value and variance to get an upper-bound estimate on computation time, i.e., $\Hat{t}_{c,n} = x_{n|n-1} + \beta p_{n|n-1}$. We choose the parameter $\beta$ accordingly to get sufficiently high confidence as an upper bound assuming a Gaussian distribution. It is important to note here that the addition of an adaptive Kalman filter for estimation of computation time doesn't significantly affect the computation time of the overall algorithm, as augmenting by a fixed number of computations in the cycle leads to an insignificant increase in computation time as opposed to window-based estimation methods whose computational complexity depends on the window size.

    

\subsection{Controller Plan A}

\begin{figure}[htbp] 
    \centering
    \includegraphics[width=0.45\textwidth]{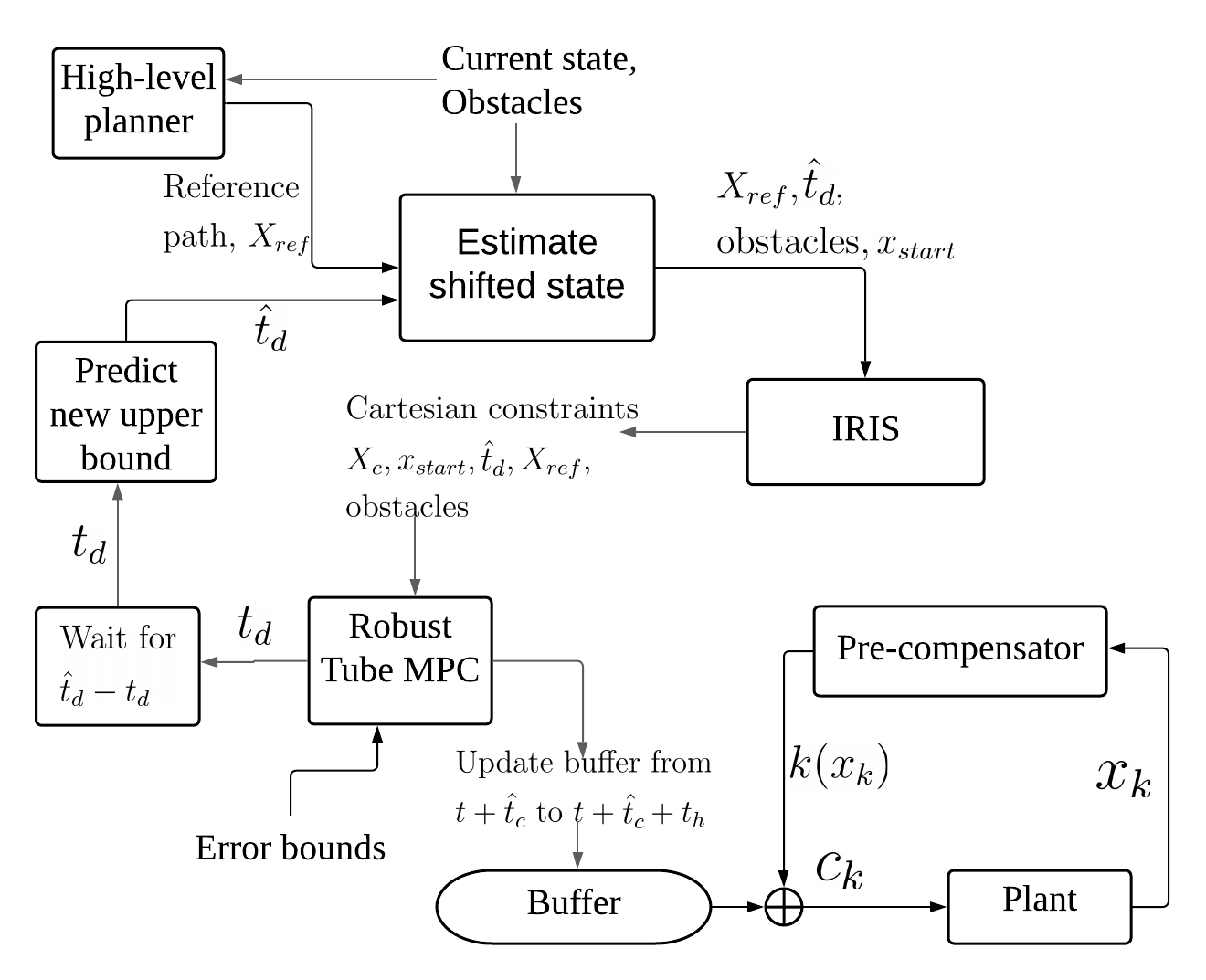}
    \caption{Controller Plan A for Robust tube MPC. If the true delay $t_d$ is smaller than the estimated $\hat{t}_d$, we need to wait $\hat{t}_d-t_d$ for a time alignment.}
    \label{planA_fig}
\end{figure}

We present a standalone controller (called plan A) as depicted in Fig. \ref{planA_fig}. We compensate for the actuator's steering dynamics by modelling a first-order ODE. For the compensation of the computation and actuator command processing delays, we use initial state shift by the estimated local upper bound on the net delay time. The optimization problem updates the robust tube buffer from $t+\hat{t}_c$ to $t+\hat{t}_c+t_h$ with the nominal commands and states. The pre-compensator unit runs as a low-level process to refine the control with a higher frequency (see Fig. \ref{planA_fig}).
\vspace{-0.4cm}
\vspace{-0.4cm}
\subsection{Controller Plan B}
\begin{figure}[htbp] 
    \centering
    \includegraphics[width=0.5\textwidth]{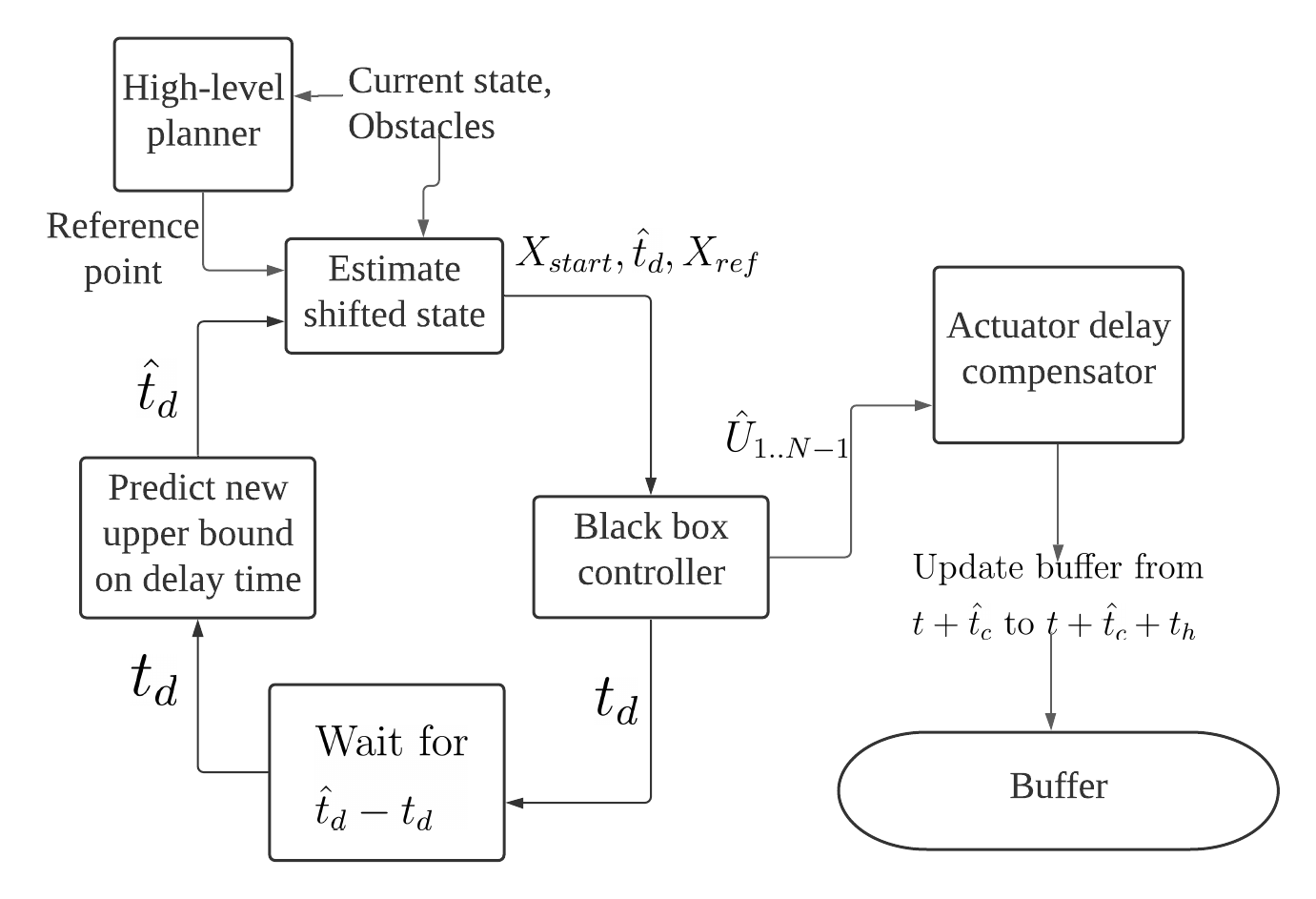}
    \caption{Controller Plan B for black box controller.}
    \label{planB_fig}
\end{figure}

As an alternate plan, we compensate for the actuator and computation delay of a nominal controller (see Fig. \ref{planB_fig}). It can be a black-box controller, which can be used for a LE controller trained in a simplified simulation environment without any practical delay. For the computation time and actuator processing delay compensation, we use the same design as plan A by shifting the initial state. For actuator dynamic delay compensation, we use a separate unit which takes the commands from the nominal controller as a reference to track. The computation time and actuator processing delays are estimated via Plan A to shift the initial state of the LE controller and conduct rollouts to obtain sequential commands $\Hat{U} = \{\Hat{u}_1,\Hat{u}_2...,\Hat{u}_N\}$. The refined commands after compensating for actuator dynamic delay are $U = \{u_1,u_2..,u_N\}$. $U$ is obtained by solving a quadratic optimization problem (Eq. \ref{actuator_dynamics_compensator}) where $u_{start}$ is the current value of the steering angle, $r_k$ is the unit step response of the steering actuator at the $k^{th}$ time step, and $Q_{ac}$ and $R_{ac}$ are positive semidefinite weight matrices. The optimization is to track the desired actuator commands from the LE controller as closely as possible while minimizing the control effort. Though collision avoidance is not considered in the experiments for Plan B, safety constraints such as control barrier functions \cite{ames2019control} can be easily incorporated into the optimization.
\vspace{-0.2cm}
\begin{equation} \label{actuator_dynamics_compensator}
    \begin{split}
    \mathop{\min}\limits_{U}&\quad\sum_{k=1}^{N} \left\Vert \Hat{u}_k - (u_0 + \sum_{i=1}^{i=k} (u_i - u_{i-1}) r_{k-i+1}) \right\Vert_{Q_{ac}} \\
    &\quad + \left\Vert u_k \right\Vert_{R_{ac}} \\
    s.t.&\quad u_0 - u_{start} = 0\\
    \end{split} 
\end{equation}
where $r_i = (1 - e^{-K i \Delta t})$.

\section{Simulation Results}
\label{sec:experiment}

\begin{figure}[htbp]
    \centering
    \includegraphics[width=0.45\textwidth]{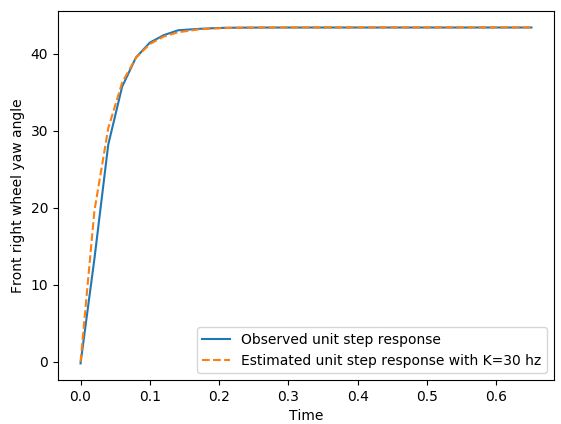}
    \caption{Steering angle response for Prius model in Gazebo}
    \label{response_fig}
\end{figure}

We conduct the experiments in the Gazebo simulator with a Prius vehicle model. In order to get the time constant value for the steering actuator, we test its unit response, i.e., we set the actuator command to 1 and record the steering angle values over a time window sufficient for the steering angle to converge at the maximum value. We then fit the observed response values with the first-order ODE described in Section \ref{actuator_dynamics} and determine the parameter $K_{\delta}$. As shown in Fig. \ref{response_fig}, using $K_{\delta} = 30 $ hz well approximates the actuator dynamics for the Prius model in Gazebo. We have also attached a short compiled video\footnote[1]{\href{https://www.youtube.com/watch?v=KdqClN4fiR0}{https://www.youtube.com/watch?v=KdqClN4fiR0}} run for all the experiments discussed.

\subsection{Static scenario} \label{sec:expA}

\begin{figure}[htbp] 
    \centering
    \includegraphics[width=0.45\textwidth]{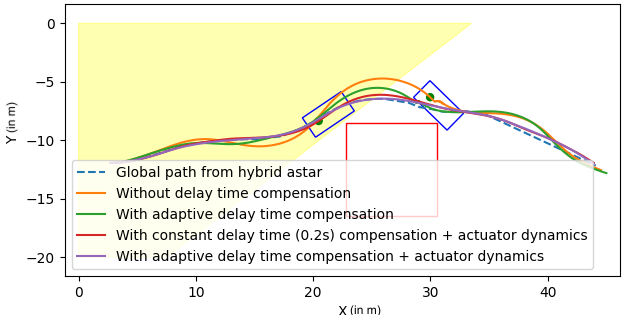}
    \caption{Comparison of paths followed. Yellow region is the convex state constraint set from IRIS at pose A for Experiment \ref{sec:expA}. The red box is a static obstacle.}
    \label{fig:fig1_paths}
\end{figure}

\begin{figure}[htbp]
    \centering
    \includegraphics[width=0.45\textwidth]{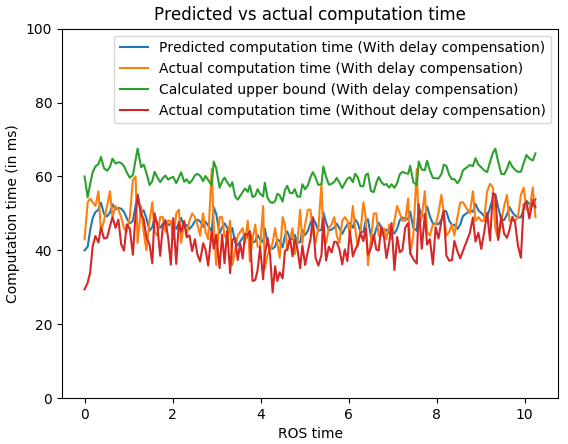}
    \caption{Observed and estimated computation times using INFLUENCE for Experiment \ref{sec:expA}.}
    \label{fig:fig1_times}
\end{figure}

For testing controller plan A, we perform a static obstacle avoidance experiment. The planner used is hybrid A$^\ast$ \cite{dolgov2008practical}. We compare the paths followed by the controller with and without considering delays. In the case in which no compensation is considered, due to the delays, the vehicle overshoots during the first turn and when it tries to get back to the reference safely, the vehicle collides with the static obstacle as marked by pose B in Fig. \ref{fig:fig1_paths}. In the cases in which delay compensations are considered, the resulting path followed is closer to the reference line and smoother, while safely avoiding collision. We also compare the results between when 1) the delay time is taken as an upper bound equal to the horizon length \cite{su2013computation} and 2) the local upper bound estimate is found using INFLUENCE. The path followed using our proposed method is clearly seen to be smoother in Fig. \ref{fig:fig1_paths}. This is because in the case of constant delay compensation, at pose A, the state constraints generated from IRIS force the vehicle to deviate from the reference path, which thus overshoots by a significant amount, but is still able to get back to the reference safely. On the other hand, if we approximate the delay time using INFLUENCE and adjust the expected local upper bound value accordingly (see Fig. \ref{fig:fig1_times}), the controller responds faster. Hence, after passing pose A, the state constraints change and the controller reacts faster to get back to the reference path, giving less overshoot. It is important to note here that the minute computation time difference between the with and without delay compensation cases is due to the shifted state calculation and the extended Kalman filter, but it seems that having an efficient delay-aware algorithm outweighs any effect the increased computation time has on making the controller less responsive.



\subsection{Overtaking scenario} \label{sec:expB}

\begin{figure}[htbp]
\begin{subfigure}{.5\textwidth}
    \centering
    \includegraphics[width=.9\textwidth]{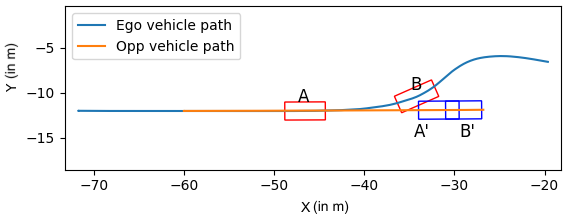}
    \caption{With variable local upper bound on computation delay.}
    \label{fig:dyn_with_comp}
\end{subfigure}
\newline
\begin{subfigure}{.5\textwidth}
    \centering
    \includegraphics[width=.9\textwidth]{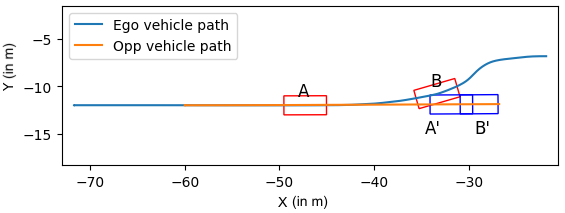}
    \caption{Without computation delay consideration.}
    \label{fig:dyn_without_comp}
\end{subfigure}
\newline
\begin{subfigure}{.5\textwidth}
    \centering
    \includegraphics[width=.9\textwidth]{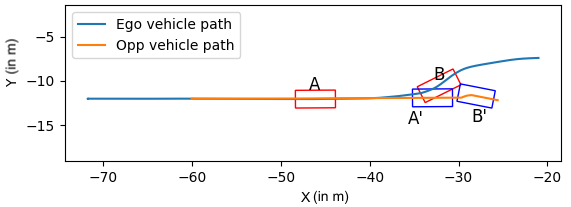}
    \caption{With constant upper bound on computation delay (0.2s).}
    \label{fig:dyn_with_comp_const}
\end{subfigure}
\caption{Comparison in Experiment \ref{sec:expB}. The red boxes denote the ego vehicle at A and B, while the blue boxes denote the opponent vehicle at $A'$ and $B'$.}
\end{figure}

We further test the controller plan A in an overtaking scenario, in which the lead vehicle brakes suddenly at point A when $t=4s$. We use the Frenet planner \cite{werling2010optimal} with the reference path as the lane center. Frenet frame-based planning has been successful in practice due to the significant advantage of its independence from complex road geometry. We perform the experiment with the same starting conditions and compare the results with (Fig. \ref{fig:dyn_with_comp}) and without (Fig. \ref{fig:dyn_without_comp}) delay compensation. The Frenet planner expects the ego vehicle to move at constant speed, but as the speed rapidly drops at point A, the reference path changes rapidly. The Frenet planner thus rapidly changes path after point A. Point B is the closest position between the ego vehicle and lead vehicle in all the cases. If delay time is not considered, the ego vehicle hits the other vehicle slightly at point B. Also, in this case if computation time is taken as a constant upper bound of $0.2s$, due to the slow reaction of the controller, the ego vehicle hits the lead vehicle at point B (Fig. \ref{fig:dyn_with_comp_const}), which proves that taking the computation time as an upper bound is ineffective in a rapidly changing environment.

\subsection{Closed track scenario} \label{sec:expC} 

\begin{figure}[htbp] 
    \centering
    \includegraphics[width=0.45\textwidth]{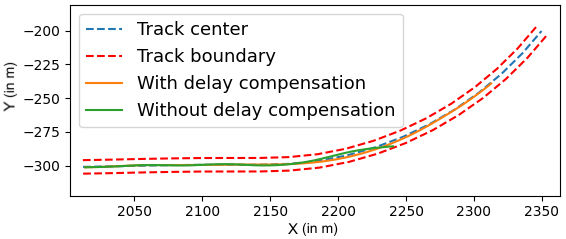}
    \caption{Comparison of high-speed tracking for Experiment \ref{sec:expC}}
    \label{fig:closed_track_fig}
\end{figure}

We test controller plan B for a LE controller trained in an ideal environment without delays. The LE lateral controller is a neural network trained on waypoint following with inputs $[\Delta x, \Delta y, \Delta \theta]$, where $(\Delta x, \Delta y)$ and $\Delta \theta$ are respectively the relative position and heading with a target waypoint. The output is the steering angle $\delta$. The network architecture is a simple feed-forward neural network with hidden layer sizes $(4,16,4)$. For longitudinal control, simple PID control is used to track a constant speed of $75 \ m/s$ throughout. The simulation has been performed in the Ansys VRX simulator \cite{vrx}. When deploying the LE controller in such a high-speed waypoint-following scenario in Gazebo, it performs worse due to the practical delays. The LE controller loses control at the time of turning, see Fig. \ref{fig:closed_track_fig}. By using the proposed plan B, the vehicle retains control. The vehicle is operating at its friction limits, hence even a little bit of error caused due to delay leads to the vehicle losing control, even when the computation time is just around 0.02s.
\section{Conclusion}
\label{sec:conclusions}

We propose a unified framework for compensating the delays of computation, actuator command processing and actuator dynamics in autonomous driving systems. We propose the INFLUENCE algorithm to safely approximate the computation time. With the use of tube MPC, the vehicle safely tracks the planned trajectories in realistic scenarios tested in the high-fidelity Gazebo simulator. Lastly, we present a framework for compensating delays for a black-box controller trained in an ideal environment. The simulation results demonstrate safety and real-time performance of our proposed framework. 

\section*{Acknowledgment}

This material is based upon work supported by the United States Air Force and DARPA under Contract No. FA8750-18-C-0092. Any opinions, findings and conclusions or recommendations expressed in this material are those of the author(s) and do not necessarily reflect the views of the United States Air Force and DARPA.

\bibliographystyle{./bibliography/IEEEtran}
\bibliography{./bibliography/IEEEabrv,./bibliography/IEEEexample}

\end{document}